\documentclass[letterpaper]{article} 
\usepackage{aaai24}  
\usepackage{times}  
\usepackage{helvet}  
\usepackage{courier}  
\usepackage[hyphens]{url}  
\usepackage{graphicx} 
\usepackage{natbib}  
\usepackage{caption} 
\usepackage{algorithm}
\usepackage{algorithmic}
\usepackage{bm}

\usepackage{newfloat}
\usepackage{listings}
\DeclareCaptionStyle{ruled}{labelfont=normalfont,labelsep=colon,strut=off} 
\floatstyle{ruled}
\newfloat{listing}{tb}{lst}{}
\floatname{listing}{Listing}
\title{Progressive Painterly Image Harmonization from Low-level Styles to High-level Styles}
\author{
    Li Niu\thanks{Corresponding author.},
    Yan Hong, 
    Junyan Cao, 
    Liqing Zhang 
    \\
}
\affiliations{
    MoE Key Lab of Artificial Intelligence, Shanghai Jiao Tong University\\

    \{ustcnewly, hy2628982280, joy\_c1, lqzhang\}@sjtu.edu.cn
}

\usepackage{bibentry}

\begin{document}

\maketitle

\begin{abstract}
Painterly image harmonization aims to harmonize a photographic foreground object on the painterly background. Different from previous auto-encoder based harmonization networks, we develop a progressive multi-stage harmonization network, which harmonizes the composite foreground from low-level styles (\emph{e.g.}, color, simple texture) to high-level styles (\emph{e.g.}, complex texture). Our network has better interpretability and harmonization performance. Moreover, we design an early-exit strategy to automatically decide the proper stage to exit, which can skip the unnecessary and even harmful late stages. Extensive experiments on the benchmark dataset demonstrate the effectiveness of our progressive harmonization network. Code and model are available at \url{https://github.com/bcmi/ProPIH-Painterly-Image-Harmonization}. 
\end{abstract}

\section{Introduction} \label{sec:intro}

\begin{figure}[t]
\centering
\includegraphics[width=0.99\linewidth]{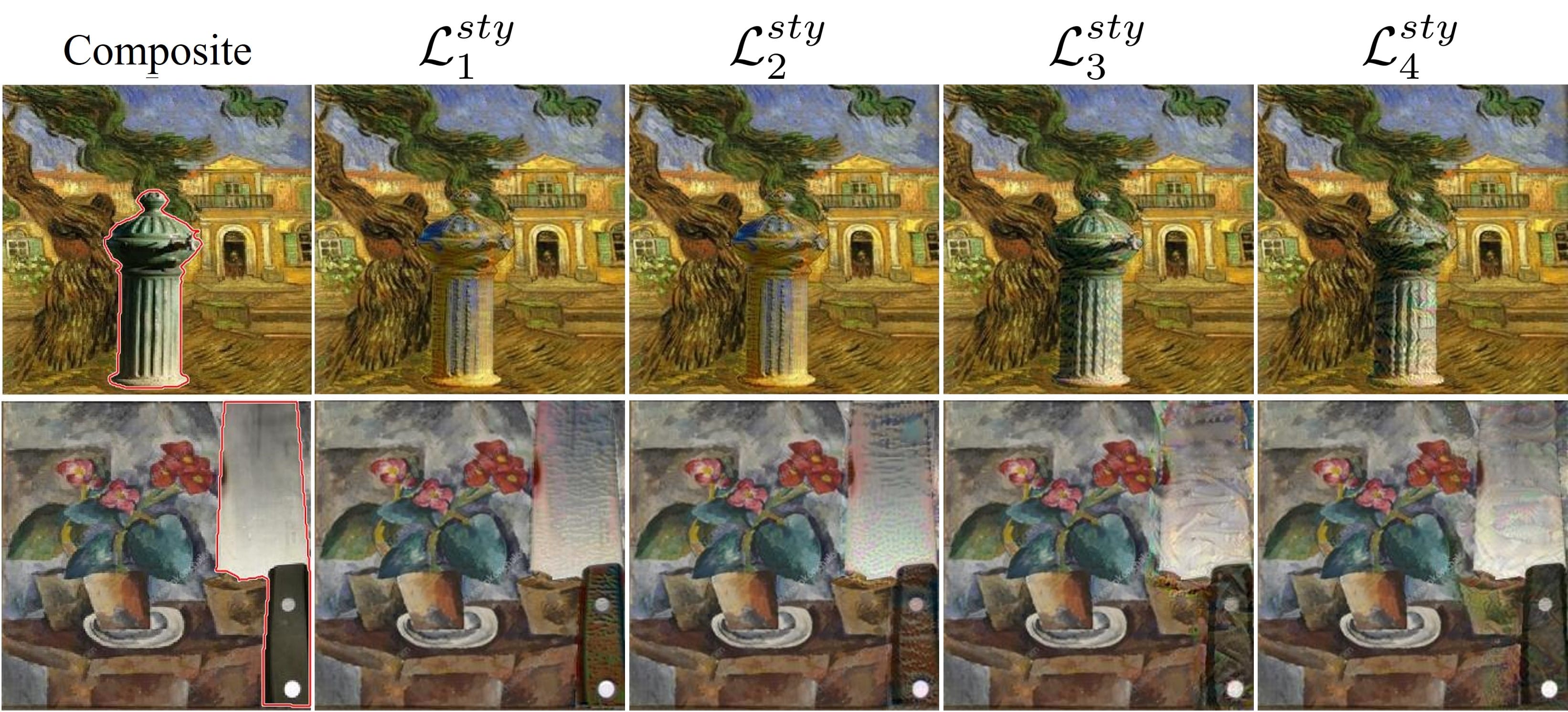}
\caption{The leftmost column shows the composite images with the foregrounds outlined in red. The rest columns show the harmonized images obtained by minimizing the style loss $\mathcal{L}^{sty}$ in different VGG-19 encoder stages, \emph{e.g.}, $\mathcal{L}^{sty}_1$ is the style loss in the first VGG-19 encoder stage.}
\label{fig:style_levels}
\end{figure}

As a common photo editing technique, image composition refers to splicing a foreground object from one image and overlaying it on another background image. However, the foreground and background in the obtained composite image may have incompatible styles, which severely harms the quality of composite image. 
Given a composite image, if the foreground is from a photographic image and the background is an artistic painting (see Figure \ref{fig:style_levels}), the background style could be defined similarly as in artistic style transfer \cite{gatys2016image,huang2017arbitrary,park2019arbitrary}, which includes color, texture, and so on. The task to mitigate the style mismatch between foreground and background in such composite images is 
called painterly image harmonization \cite{luan2018deep}, which adjusts the style of foreground to make it compatible with the background and naturally embedded into the background. 

As far as we are concerned, there exist only few works on painterly image harmonization, which can be divided into optimization-based methods \cite{luan2018deep,zhang2020deep} and feed-forward methods \cite{cao2022painterly,yan2022style}. Optimization-based methods \cite{luan2018deep,zhang2020deep} update the composite foreground iteratively to minimize the designed loss functions, which is too slow for real-time application. In contrast, feed-forward methods only pass the composite image through the network once to produce a harmonized image, which is much more efficient than optimization-based methods. Among the feed-forward methods, \citet{peng2019element} introduced AdaIN \cite{huang2017arbitrary} into painterly image harmonization. \citet{yan2022style} proposed to combine the advantages of transformer and CNN. \citet{cao2022painterly} explored harmonizing the composite image in both spatial domain and frequency domain. Some diffusion model based methods~\cite{sdedit,cdc,TFICON} for cross-domain image composition can also be applied to this task. However, the above methods are still struggling to preserve the content details and transfer the styles sufficiently. Moreover, the above methods directly output the final harmonized image, which function like a black box. 

In this work, we propose an interpretable network structure which harmonizes the composite image progressively. In previous works \cite{peng2019element,cao2022painterly}, style is usually represented by the feature statistics (\emph{e.g.}, mean, variance) of pretrained VGG-19 \cite{VGG19} encoder. We observe that the feature statistics in shallow and deep encoder stages represent low-level and high-level styles respectively, which has also been discussed in previous literature \cite{yao2019attention,cho2019image,liu2021adaattn}.
As shown in Figure \ref{fig:style_levels}, we optimize the composite foreground by minimizing the style loss in different encoder stages and the content loss (see Section \ref{sec:main_network} for the details of style and content loss). From left to right, the adjusted style of composite foreground transits from low-level style (\emph{e.g.}, color, simple texture) to high-level style (\emph{e.g.}, complex texture). In the first row, $\mathcal{L}^{sty}_1$ changes the foreground color from green to yellow, while $\mathcal{L}^{sty}_4$ adds wave texture to the foreground.  In the second row,  $\mathcal{L}^{sty}_1$ changes the small-scale texture of foreground, while $\mathcal{L}^{sty}_4$ changes the large-scale texture, in which the large-scale texture means that the texture is perceivable with large receptive field. 

Motivated by these observations, we design a dual-branch network which harmonizes the composite image from low-level styles to high-level styles progressively. The top branch is a pretrained VGG-19 \cite{VGG19} encoder with four encoder stages, in which we apply AdaIN \cite{huang2017arbitrary} to harmonize the output feature map in each encoder stage. The bottom branch has four stages corresponding to four encoder stages in the top branch, and maintain the full resolution through four stages.  In the $k$-th stage, the bottom branch fuses the harmonized feature maps from the first to the $k$-th encoder stage in the top branch, and the output image is harmonized up to the $k$-th style level. \emph{The progressive network has several advantages over previous networks that entangle all levels of styles.} Firstly, our network owns better interpretability, reflecting how the composite image is harmonized from low-level styles to high-level styles. Secondly, from low-level style transfer to high-level style transfer, the multi-stage network gradually raises the difficulty level of harmonization task, which might be easier than handling all style levels simultaneously.

In practice, we observe that for some composite images, the early stages could produce comparable or even better results than the late stages.  One reason is that in some cases, it is sufficient to only transfer low-level styles. Another reason is that when struggling to transfer high-level styles, the content information may be sacrificed.  Therefore, we further propose an early-exit strategy, which uses sequential model  to forecast the proper stage to exit the network. \emph{The early-exit strategy can benefit the harmonization quality by discarding unnecessary and even harmful late stages.} We name our method as ProPIH (\textbf{Pro}gressive \textbf{P}ainterly \textbf{I}mage \textbf{H}armonization).

Our contributions can be summarized as follows: 1) We design a progressive painterly harmonization network ProPIH, which can harmonize the composite image from low-level styles to high-level styles. 2) We propose an early-exit strategy, which can automatically determine the proper stage to exit the network. 3) Extensive experiments on COCO \cite{lin2014microsoft} and WikiArt \cite{nichol2016painter} demonstrate the effectiveness of our proposed method.

\section{Related Work}

\subsection{Image Harmonization} \label{sec:image_harmonization}

Image harmonization aims to harmonize a composite image by adjusting foreground illumination to match background illumination. 
In recent years, abundant deep image harmonization methods \cite{tsai2017deep,Jiang_2021_ICCV,xing2022composite,peng2022frih,zhu2022image,valanarasu2022interactive,LEMaRT} have been developed. For example, \cite{xiaodong2019improving,Hao2020bmcv,sofiiuk2021foreground} proposed diverse attention modules to treat the foreground and background separately, or establish the relation between foreground and background. \cite{cong2020dovenet,ling2021region,hang2022scs} directed image harmonization to domain translation or style transfer by treating different illumination conditions as different domains or styles.  \cite{guo2021image,guo2021intrinsic,guo2022transformer}  decomposed an image into reflectance map and illumination map.  More recently, \cite{cong2022high,ke2022harmonizer,liang2021spatial,xue2022dccf,PCTNet,WangCVPR2023} used deep network to predict color transformation, striking a good balance between efficiency and effectiveness. 
However, most image harmonization methods only adjust illumination and require ground-truth supervision, which is unsuitable for painterly image harmonization.
\subsection{Painterly Image Harmonization}

Different from Section \ref{sec:image_harmonization}, in painterly image harmonization, the foreground is a photographic object while the background is an artistic painting. The goal of painterly image harmonization is adjusting the foreground style to match background style and preserving the foreground content. The existing methods can be roughly categorized into optimization-based methods and feed-forward methods.  The optimization-based methods \cite{luan2018deep,zhang2020deep} iteratively optimize over the composite foreground to minimize the designed loss functions. The feed-forward methods \cite{peng2019element,yan2022style,cao2022painterly} send the composite image through the harmonization network once and generate the harmonized image. 
Several diffusion model based methods~\cite{sdedit,cdc} for cross-domain image composition can also be used to harmonize the composite image. 

Different from previous network structures, we design a novel progressive harmonization network which can harmonize a composite image from low-level to high-level styles.

\subsection{Artistic Style Transfer}

Artistic style transfer~\cite{kolkin2019style,jing2020dynamic,chen2021dualast,chen2017stylebank,sanakoyeu2018style,wang2020collaborative,li2018learning,chen2016fast,sheng2018avatar,gu2018arbitrary,zhang2019multimodal,chen2022toward,huo2021manifold} targets at recomposing a content image in the style of a style image. Amounts of works \cite{huang2017arbitrary,li2017universal,li2019learning} focus on global style transfer. 
Recently, some works \cite{park2019arbitrary,liu2021adaattn,deng2022stytr2} turn to fine-grained local style transfer by establishing local correspondences between content image and style image. More recently, diffusion model has been explored for artistic style transfer~\cite{inst}.  Artistic style transfer stylizes an entire content image, whereas painterly image harmonization stylizes the foreground object in the composite image.

A few style transfer methods ~\cite{wang2017multimodal,luo2022progressive} have coarse-to-fine network structure, but \emph{``coarse-to-fine" means moving from coarse-grained structure to fine-grained details, which is essentially different from our transition from low-level styles to high-level styles.} 
Some style transfer works \cite{yao2019attention,cho2019image,liu2021adaattn} have discussed low-level and high-level styles, but they still transfer multi-level styles simultaneously, instead of from low-level styles to high-level styles progressively.

\section{Our Method}

\begin{figure}[t]
\centering
\includegraphics[width=0.99\linewidth]{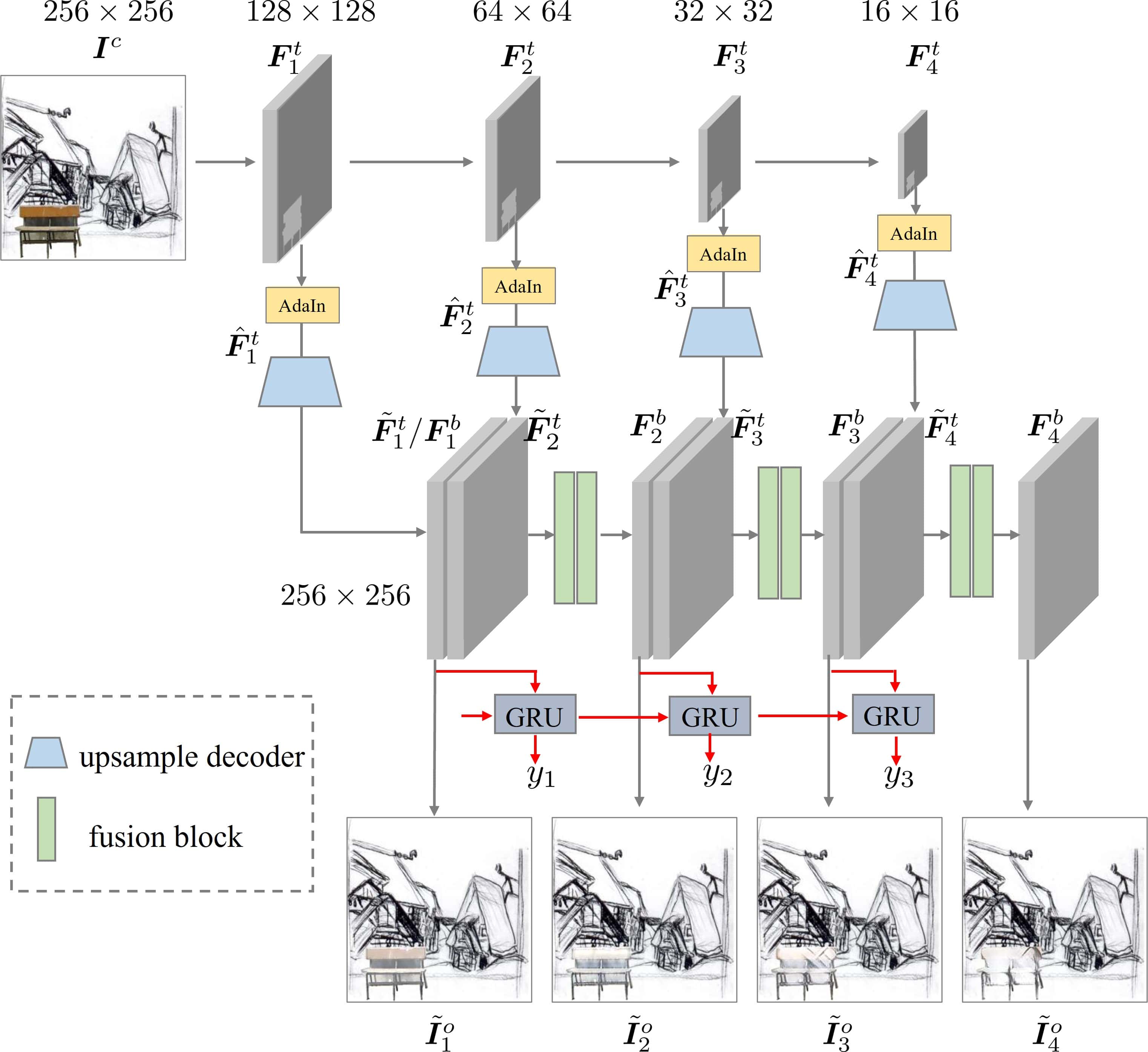}
\caption{Our ProPIH consists of the top branch (pretrained VGG-19 encoder) and the bottom branch. The $k$-stage in the bottom branch fuses the encoder features up to the $k$-th stage, and produces the image harmonized up to the $k$-th style level. GRU predicts the exit label indicating whether to exit the network in the current stage.}
\label{fig:network}
\end{figure}

We suppose that a composite image $\bm{I}^c$ is obtained by pasting a photographic foreground on the painterly background $\bm{I}^b$. We use foreground mask $\bm{M}^f$ to indicate the foreground region and $\bm{M}^b$ is the complementary background mask. Painterly image harmonization aims to transfer the style from background region to foreground region while preserving the content of foreground, giving rise to a harmonized image $\tilde{\bm{I}}^o$.

As shown in Figure~\ref{fig:network}, our network has two branches. The top branch is the pretrained VGG-19 network~\cite{VGG19}. The bottom branch progressively fuses the upsampled encoder feature maps from the top branch to produce the multi-stage outputs which are harmonized up to different style levels. Specifically, the output from early (\emph{resp.}, late) stage is harmonized up to low (\emph{resp.}, high) style level. 
We also attach GRU to the bottom branch to predict the exit stage, so that we can exit the network at the earliest proper stage. Next, we will introduce our network structure in Section~\ref{sec:main_network} and early-exit strategy in Section~\ref{sec:early_exit}.

\subsection {Progressive Harmonization Network} \label{sec:main_network}

As illustrated in Figure~\ref{fig:network}, the top branch is the pretrained VGG-19 encoder~\cite{VGG19}, where we only use the first few layers up to \emph{ReLU-4\_1}. We divide the encoder to four stages, and the last layer of four stages are \emph{ReLU-1\_1}, \emph{ReLU-2\_1}, \emph{ReLU-3\_1}, and \emph{ReLU-4\_1} respectively. The output feature map from the $k$-th stage in the top branch is denoted as $\bm{F}^{t}_k$, which is comprised of the foreground feature map  $\bm{F}^{t,f}_k$ and background feature map  $\bm{F}^{t,b}_k$. To transfer the background style to the foreground, we match the channel-wise statistics (mean, standard deviation) of $\bm{F}^{t,f}_k$ with those of $\bm{F}^{t,b}_k$ using Adaptive Instance Normalization (AdaIN) \cite{huang2017arbitrary}, which can be formulated as follows,
\begin{eqnarray}\label{eqn:adain} 
    \hat{\bm{F}}^{t,f}_{k} = \sigma(\bm{F}^{t,b}_k)\frac{\bm{F}^{t,f}_k-\mu(\bm{F}^{t,f}_k)}{\sigma(\bm{F}^{t,f}_k)} + \mu(\bm{F}^{t,b}_k),
\end{eqnarray}
in which $\mu(\cdot)$ (\emph{resp.}, $\sigma(\cdot)$) means calculating the mean (\emph{resp.}, standard deviation) of the specified feature map. The stylized foreground $\hat{\bm{F}}^{t,f}_{k}$ is combined with the unchanged background $\bm{F}^{t,b}_k$ to form the harmonized feature map $\hat{\bm{F}}^{t}_{k}$.  

For each stage in the top branch, we attach a lightweight decoder which upsamples the harmonized feature map to the full resolution (\emph{e.g.}, $256\times 256$). The upsample decoder for the $k$-th stage contains one conv block and $k$ upsample blocks. We denote the output feature map from the $k$-th upsample decoder as $\tilde{\bm{F}}^t_k$.  

In the bottom branch, for the first stage, we directly use the output feature map $\tilde{\bm{F}}^t_1$ to produce the harmonized image $\tilde{\bm{I}}_1^o$ through a $3\times 3 $ conv layer. We use $\bm{F}^{b}_k$ to represent the final feature map of the $k$-th stage in the bottom branch, so $\bm{F}^{b}_1=\tilde{\bm{F}}^t_1$ in the first stage.
For the rest of stages, we adopt two fusion blocks to fuse the final feature map in previous stage and the output from the upsample decoder. Specifically, for the $k$-th stage ($k>1$), we concatenate $\bm{F}^{b}_{k-1}$ with $\tilde{\bm{F}}^t_k$, which passes through two fusion blocks to arrive at $\bm{F}^{b}_k$. $\bm{F}^{b}_k$ is used to produce the harmonized image  $\tilde{\bm{I}}^o_k$ through a $3\times 3 $ conv layer. The details of conv/upsample block in the upsample decoder and fusion block in the bottom branch can be found in Section \ref{sec:imp_detail}.

The $k$-th stage in the bottom branch integrates the information of the harmonized feature maps up to the $k$-th stage in the top branch, that is, $\{\hat{\bm{F}}^{t}_1,\ldots, \hat{\bm{F}}^{t}_k\}$. Accordingly, we enforce the output image $\tilde{\bm{I}}^o_k$ in the $k$-th stage to be harmonized up to the $k$-th style level, so the style loss \cite{huang2017arbitrary} for  $\tilde{\bm{I}}^o_k$ can be written as
\begin{eqnarray}\label{eqn:style_loss}
\mathcal{L}^{sty}_k=\!\!\!\!\!\!\!\!\!\!\!\!&&\sum_{k'=1}^{k}\|\mu\left(\phi_{k'}(\tilde{\bm{I}}^{o}_k)\circ \bm{M}^f\right)-\mu(\hat{\bm{F}}^{t,f}_{k'}) \|^2 \nonumber\\
&&\!\!\!\!\!\!\!\!\!\!\!\!+ \sum_{k'=1}^{k}\|\sigma\left(\phi_{k'}(\tilde{\bm{I}}^{o}_k)\circ \bm{M}^f\right)-\sigma(\hat{\bm{F}}^{t,f}_{k'}) \|^2, 
\end{eqnarray}
where each $\phi_{k'}(\cdot), k'\in\left\{1,2,3,4\right\}$ denotes the output feature map from the $k'$-th VGG-19 encoder stage, and $\circ$ means element-wise product. $\mu(\cdot)$ and $\sigma(\cdot)$ are the same as defined in Eqn.~\ref{eqn:adain}.

Additionally, for the harmonized images of all stages, we utilize the same content loss \cite{gatys2016image} to ensure that the foreground content is maintained:
\begin{equation}\label{eqn:content_loss}
    \mathcal{L}^{con}_k =\left\|\phi_4(\tilde{\bm{I}}^o_k)-\phi_4(\bm{I}^c)\right\|^2,
\end{equation}
in which $\phi_{4}(\cdot)$ is the same as defined in Eqn.~\ref{eqn:style_loss}.
For the $k$-th stage, the total loss is given by $\mathcal{L}^{total}_k = \mathcal{L}^{con}_k + \mathcal{L}^{sty}_k$.

From the first stage to the last stage, the bottom branch gradually harmonizes the composite foreground from low-level styles to high-level styles. Specifically, we first adjust the low-level styles (\emph{e.g.}, color, simple pattern) of foreground and then adjust the high-level styles (\emph{e.g.}, complex pattern) of foreground. As low-level styles could be easily transferred by only using shallow features, low-level style transfer is a simpler task than high-level style transfer. Therefore, the bottom branch transits from easy task to hard task, which might be easier than handling all tasks simultaneously, as discussed in the realm of curriculum learning \cite{bengio2009curriculum}. Moreover, during the transition from easy task to hard task, we can terminate at the proper stage to prevent the adverse effect brought by challenging the hard task, which will be introduced in Section~\ref{sec:early_exit}.

\subsection {Early-exit Strategy} \label{sec:early_exit}

Based on the harmonized results from different stages, we observe that in some cases, the harmonized results of early stages are comparable or even better than those of late stages. One reason is that it is sufficient to only transfer low-level styles for certain composite images. Another reason is that the network is sometimes struggling to transfer high-level styles at the cost of sacrificing the content information, so that the content structure of harmonized image is severely distorted. Motivated by the above observations, we design an early-exit strategy, which enables the network to terminate at the proper stage. Some previous works \cite{huangmulti,teerapittayanon2016branchynet,bolukbasi2017adaptive,mcgill2017deciding,jie2019anytime,li2019improved,figurnov2017spatially,leroux2018iamnn} on dynamic routing explored  exiting the network early. We apply similar idea to our progressive harmonization network, but the motivation and technical approach are considerably different from theirs.    

To provide effective supervision for when to exit, we manually annotate the exit stages for $5000$ randomly selected composite images in the training set (see details in the supplementary), in which the exit stage means the earliest stage that could produce satisfactory result. Given a composite image with annotated exit stage $\hat{k}$, the exit labels of stages prior to the exit stage are $0$, that is, $y_{k'}=0$ for $k'<\hat{k}$,  otherwise $y_{k'}=1$. 

We adopt sequential model to predict the exit label for each stage sequentially. For simplicity, we choose GRU~\cite{cho2014properties} as the sequential model. At the $k$-th stage in the bottom branch, we perform global average pooling over the final feature map $\bm{F}^b_k$ and produce $\bm{f}^b_k$. Then, we send $\bm{f}^b_k$ and the hidden state $\bm{h}_{k-1}$ of GRU at the $(k\!-\!1)$-th stage to the GRU cell. Formally, the function of GRU cell at the $k$-th stage can be represented by
\begin{eqnarray}
[\tilde{y}_k, \bm{h}_k] = GRU(\bm{h}_{k-1}, \bm{f}^b_k). 
\end{eqnarray}

In practice, we only add GRU cells to the first three stages and employ the binary cross-entropy loss $\mathcal{L}^{bce}_k(y_k, \tilde{y}_k)$ for the $k$-th stage. Thus, the total loss function can be written as
\begin{eqnarray}
\mathcal{L}^{all} = \sum_{k=1}^4 \mathcal{L}_k^{total} + \sum_{k=1}^3 \mathcal{L}^{bce}_k.
\end{eqnarray}

During inference, we choose the earliest stage with predicted exit score larger than the threshold $0.5$ as the exit stage. If the predicted exit scores of the first three stages are all smaller than the threshold, we proceed till the last stage.

\section{Experiments}

\subsection{Datasets and Implementation Details} \label{sec:imp_detail}

Following previous works~\cite{peng2019element,cao2022painterly}, we conduct experiments on COCO \cite{lin2014microsoft} and WikiArt \cite{nichol2016painter}.
COCO \cite{lin2014microsoft} contains instance segmentation annotations for 80 object categories, while WikiArt \cite{nichol2016painter} contains digital artistic paintings from different styles.
We create composite images based on these two datasets, with the photographic objects from COCO and the painterly backgrounds from WikiArt.
In particular, we randomly select a photographic object from COCO with the foreground ratio in $[0.05, 0.3]$. Then, we paste this photographic object on a randomly selected painterly background from WikiArt, resulting in an inharmonious composite image. 

Our model is implemented by PyTorch $1.10.0$, which is distributed on ubuntu 20.04 LTS operation system, with 128GB memory, Intel(R) Xeon(R) Silver 4116 CPU, and one GeForce RTX 3090 GPU.
We adopt Adam \cite{kingma2015adam} with  learning rate of 0.0001 as the optimization solver.  We resize the input images as $256 \times 256$ and set the batch size as $4$ for model training.
Each upsample decoder consists of one conv block and several upsample blocks. The conv block has a $3\times 3$ conv followed by ReLU. Each upsample block has an upsampling layer and a $3\times 3$ conv followed by ReLU. The conv block and upsample blocks all reduce the channel dimension by half. 
In the bottom branch, each fusion block has a sequence of $3\times 3$ conv, ReLU, $3\times 3$ conv, ReLU. The first $3\times 3$ conv in the first fusion block in each stage reduces the channel dimension by half. 

\begin{figure}[t]
\centering
\includegraphics[width=0.99\linewidth]{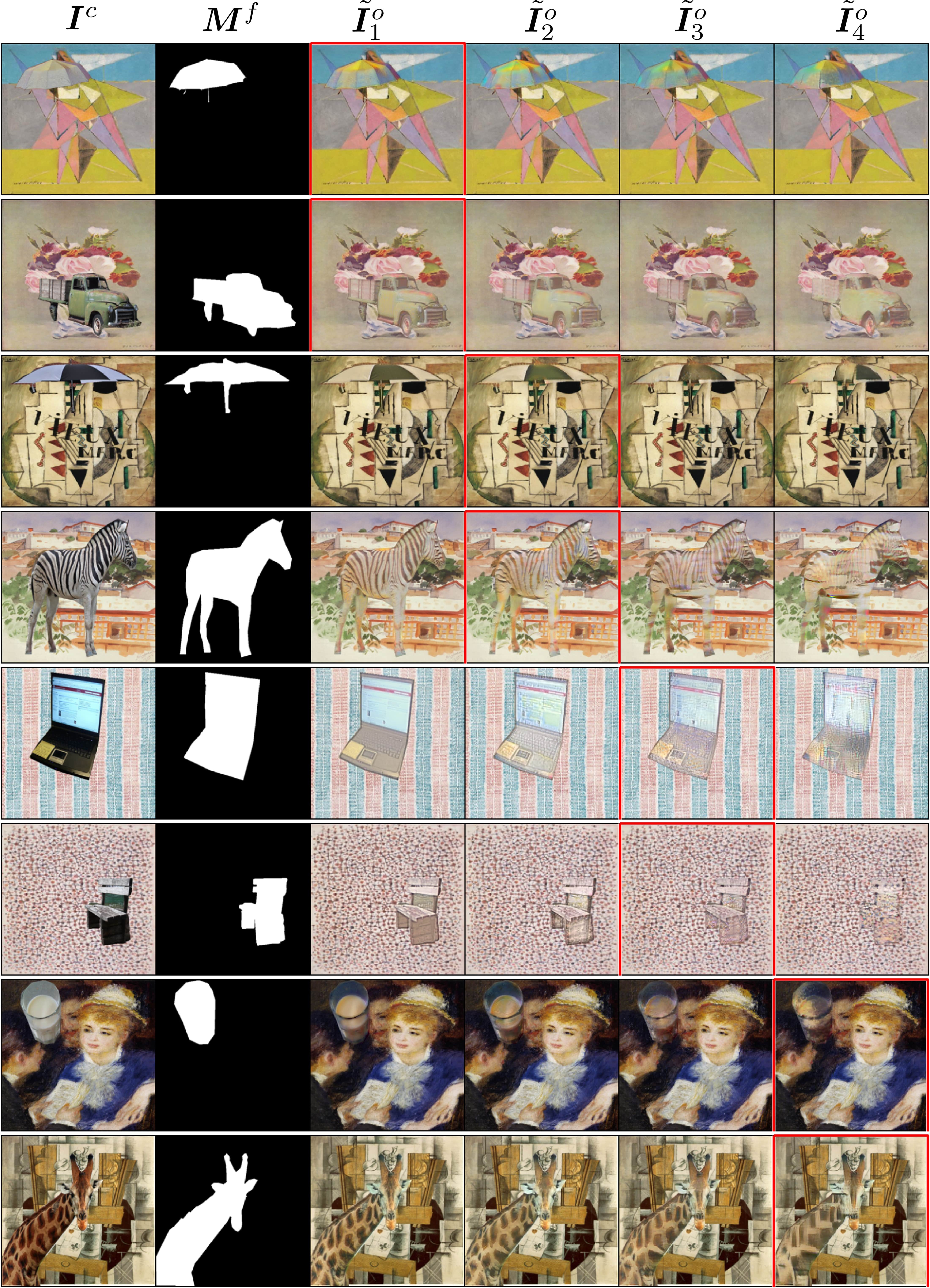}
\caption{From left to right, we show the composite image, composite mask, the harmonized results from four stages in the test set. The predicted exit stages are marked with red bounding boxes.}
\label{fig:exit_stages_test}
\end{figure}

\begin{table}[t] 
\begin{center}
\begin{tabular}{c|c|c|c}
\hline
Method  & B-T score & Time(s)  & FLOPs(G) \\
\hline
SANet & -0.429 & 0.0097 & 43.32\\
AdaAttN  & -0.626 & 0.0115 & 49.64\\
StyTr2  & 0.225 & 0.0504 & 39.74\\
QuantArt  & 0.336 & 0.1031 & 133.34 \\
Inst  & -0.762 & 2.2996 & 3378.43 \\
\hline
SDEdit & -0.654 & 2.1321 & 3164.52\\
CDC  &  0.204 & 2.3427 & 3299.81 \\
E2STN  & -0.188 & 0.0079 & 29.28\\
DPH  & 0.311 & 55.24 & -\\
PHDNet  & 0.474 & 0.0321 & 158.41\\
\hline
ProPIH  & 0.917 & 0.0067 & 23.45\\
\hline
\end{tabular}
\end{center}
\caption{The comparison between different methods.}
\label{tab:results}
\end{table}

\begin{figure*}[t]
\centering
\includegraphics[width=0.96\linewidth]{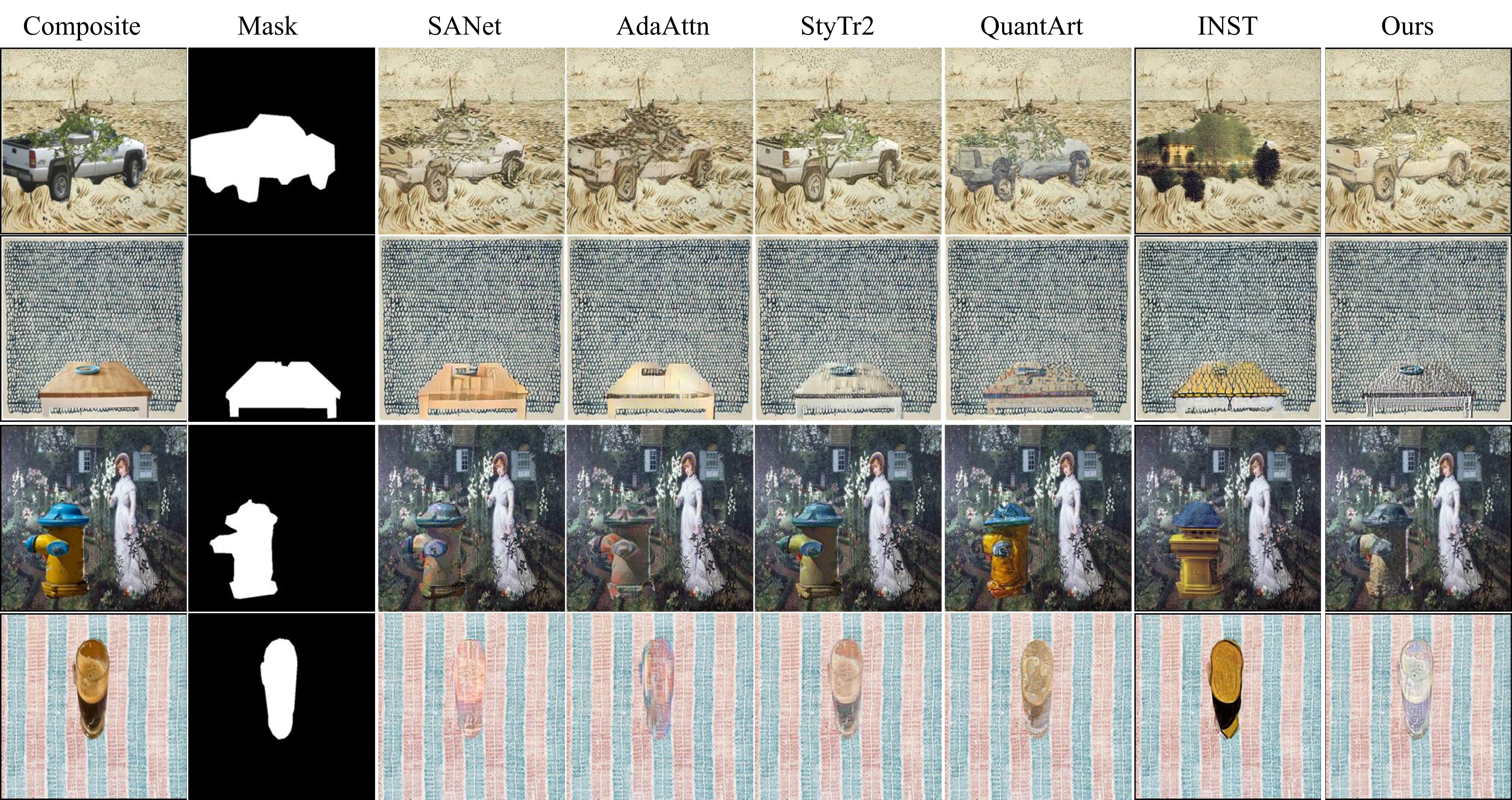}
\caption{From left to right, we show the composite image, composite foreground mask, the harmonized results of SANet~\cite{park2019arbitrary}, AdaAttN~\cite{liu2021adaattn}, StyTr2~\cite{deng2022stytr2}, QuantArt~\cite{quantart}, INST~\cite{inst}, and our ProPIH.}
\label{fig:style_transfer_baseline_main}
\end{figure*}

\begin{figure*}[t]
\centering
\includegraphics[width=0.96\linewidth]{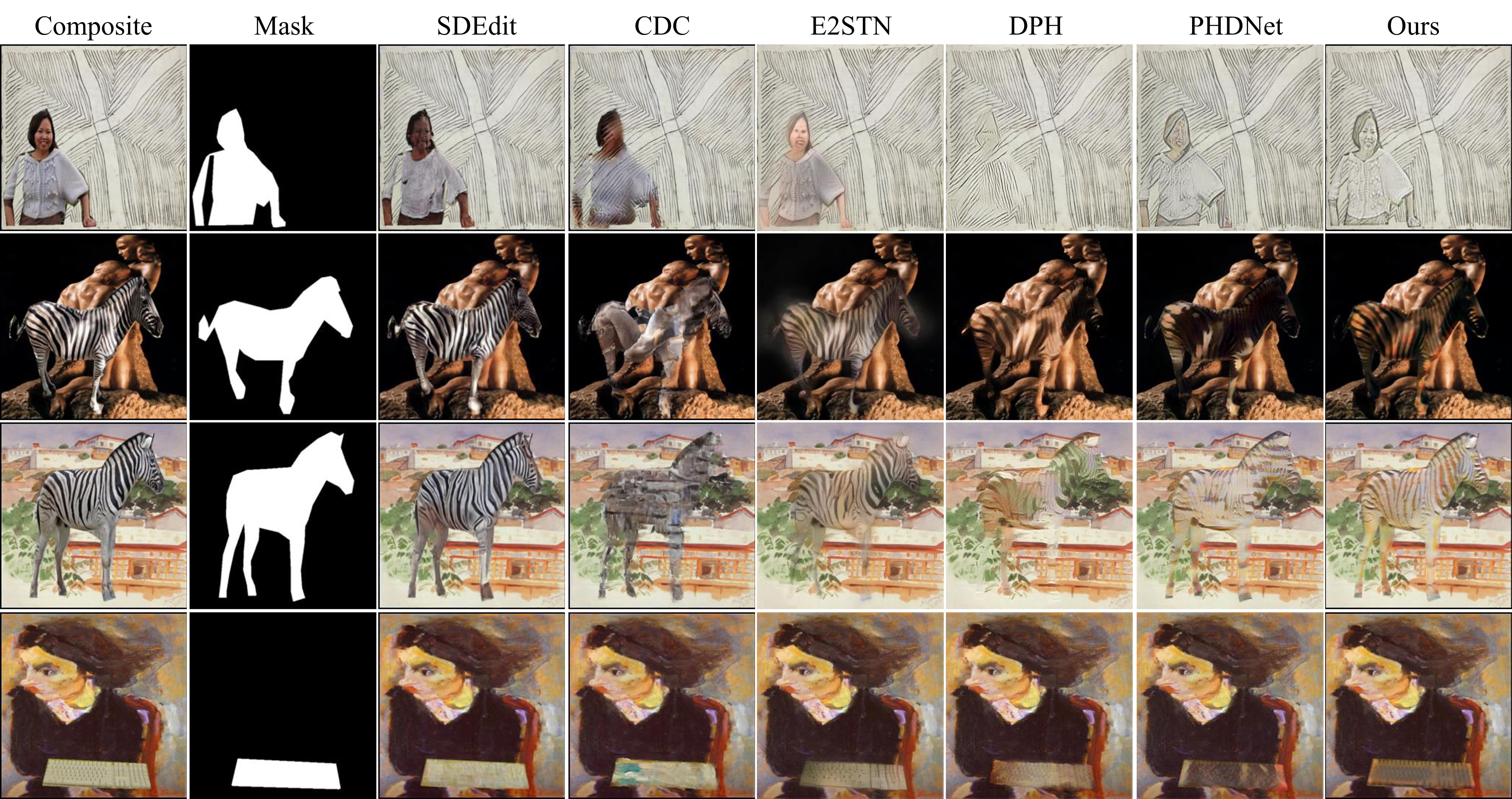}
\caption{From left to right, we show the composite image, composite foreground mask, the harmonized results of SDEdit~\cite{sdedit}, CDC~\cite{cdc}, E2STN~\cite{peng2019element}, DPH~\cite{luan2018deep}, PHDNet~\cite{cao2022painterly}, and our ProPIH.}
\label{fig:painterly_harmonization_baseline_main}
\end{figure*}

\subsection{Multiple Stages and Exit Stages} \label{sec:multi_stage}

We annotate the exit stages for $5000$ randomly selected composite images in the training set. For each composite image, we produce the results of four stages and ask human annotators to annotate its exit stage (see details in the Supplementary). 
The percentages that four stages are annotated as the exit stage are $2.1\%$, $22.3\%$, $47.4\%$, and $28.2\%$ respectively. Overall, the outputs from late stages (\emph{e.g.}, 3, 4) are better than those from early stages (\emph{e.g.}, 1, 2), because the former ones are prone to have incompatible foreground and background \emph{w.r.t.} high-level styles. Some annotation examples are provided in the Supplementary. 

We train GRU based on the annotated training images, which is then applied to the test set.
Given a composite image in the test set, its exit stage is the earliest stage with the predicted exit score larger than the threshold $0.5$. If the predicted exit scores in the first three stages are all smaller than $0.5$, the last stage is deemed as the exit stage.

We show the harmonized results from four stages and the predicted exit stages in Figure \ref{fig:exit_stages_test}. It can be seen that the outputs from early stages are harmonized up to low style level (\emph{e.g.}, color, simple pattern) while the outputs from late stages are harmonized up to high style level (\emph{e.g.}, complex pattern).
For example, in row 7-8, the outputs from the last stage have better visualization effect with more compatible foreground and background, \emph{e.g.}, fluffy glass in row 7 and the giraffe with square spots in row 8.

However, in some cases, the early stages are predicted as the exit stages. As explained in Section~\ref{sec:early_exit},  one reason is that it is sufficient to only transfer low-level styles for certain composite images. For example, in row 1-2, the outputs from the first stage have been sufficiently harmonized, probably because the background style is simple or the foreground naturally matches the background. Hence, there is no need to go through the rest of stages.  
Another reason is that the network is sometimes struggling to transfer high-level styles at the cost of sacrificing the content information, so that the content structure of harmonized image is distorted. 
For example, in row 3-6, the results from the last stage have undesired artifacts (the uneven black spots on the umbrella in row 3) or distorted content structure (the corrupted zebra strips in row 4, the distorted laptop/chair structure in row 5, 6), so we need to exit the network before the last stage, to balance between stylization and content preservation.

\subsection{Comparison with Baselines}
We compare with two groups of baselines: artistic style transfer methods and painterly image harmonization methods.  Since the standard image harmonization methods introduced in Section~\ref{sec:image_harmonization} only adjust the illumination statistics and require ground-truth images as supervision, they are not suitable for painterly image harmonization. 

For the first group, we first use artistic style transfer methods to stylize the entire photographic image according to the painterly background, and then paste the stylized photographic object on the painterly background.
We compare with typical or recent style transfer methods: SANet~\cite{park2019arbitrary}, AdaAttN~\cite{liu2021adaattn},  StyTr2~\cite{deng2022stytr2}, QuantArt~\cite{quantart}, INST~\cite{inst}. 
For the second group, we compare with SDEdit~\cite{sdedit}, CDC~\cite{cdc}, E2STN~\cite{peng2019element}, DPH~\cite{luan2018deep}, PHDNet~\cite{cao2022painterly}. 

For our method, we use the harmonized images from the predicted exit stage for comparison. 

\textbf{Visualization Results: }The comparison with the first group of baselines is illustrated in Figure~\ref{fig:style_transfer_baseline_main}. It can be seen that the baselines often fail to harmonize the composite foreground sufficiently, especially when the background has complicated texture (\emph{e.g.}, row 2). In contrast, our method could handle diversified background styles and produce visually appealing harmonized images.  

The comparison with the second group of baselines is illustrated in Figure~\ref{fig:painterly_harmonization_baseline_main}. Again, our method can produce the harmonized foreground, which has more compatible color and texture with the background. In row 1, the harmonized foregrounds of baselines have inharmonious color or distorted content structure. In row 2, the harmonized foregrounds of baselines have severe artifacts, while our method can roughly maintain the zebra strips after harmonization. In row 3 and row 4, our method can achieve a good balance between content preservation and style migration, while baselines either fail in style migration or severely distort the content structure.

\textbf{User Study: }Following \cite{cao2022painterly}, we also conduct user study to compare different methods. We randomly select 100 content images from COCO and 100 style images from WikiArt to generate 100 composite images for user study. We compare the harmonized results generated by 10 baselines and our method.

Specifically, for each composite image, we obtain $11$ harmonized outputs and use every two images from these $11$ images to construct image pairs. With $100$ composite images, we construct $5500$ image pairs. Then, we invite $100$ participants to watch one image pair each time and choose the better one. Finally, we collect $550,000$ pairwise results and calculate the overall ranking of all methods using Bradley-Terry (B-T) model~\cite{bradley1952rank,lai2016comparative}.  The results are summarized in Table~\ref{tab:results}, in which our method achieves the highest B-T score and once again outperforms other baselines.

\textbf{Efficiency Analyses: } For efficiency comparison, we report the inference time and FLOPs. We evaluate with the image size $256 \times 256$ and the inference time is averaged over 100 test images on a single RTX 3090 GPU.
Note that DPH is optimization-based method, so we omit its FLOPs.
For our method, different test images may exit the network from different stages, we count the time cost and computational cost up to the predicted exit stage for each test image.

The results are reported in Table~\ref{tab:results}. The optimization-based method DPH is time-consuming, due to iteratively updating the input composite image. Diffusion-based methods are much slower than the other feed-forward methods, which limits their real-world applicability. 
Our method is relatively efficient, because of our lightweight network structure and early-exit strategy.

\begin{figure}[t]
\centering
\includegraphics[width=0.99\linewidth]{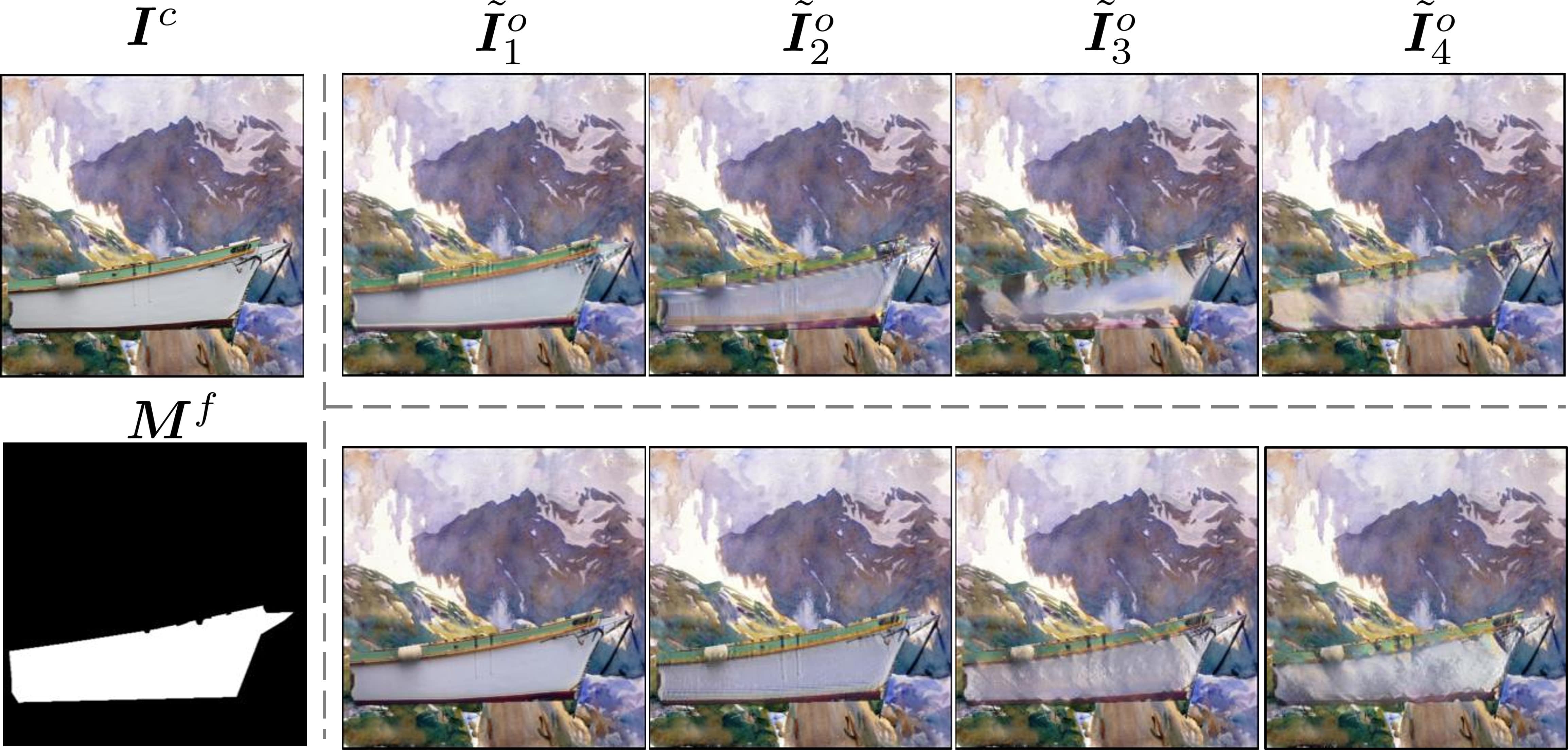}
\caption{The top row shows the harmonized results from four stages when using the full style loss for all stages. The bottom row shows the results of our default method for comparison. }
\label{fig:full_style_loss}
\end{figure}

\begin{figure}[t]
\centering
\includegraphics[width=0.99\linewidth]{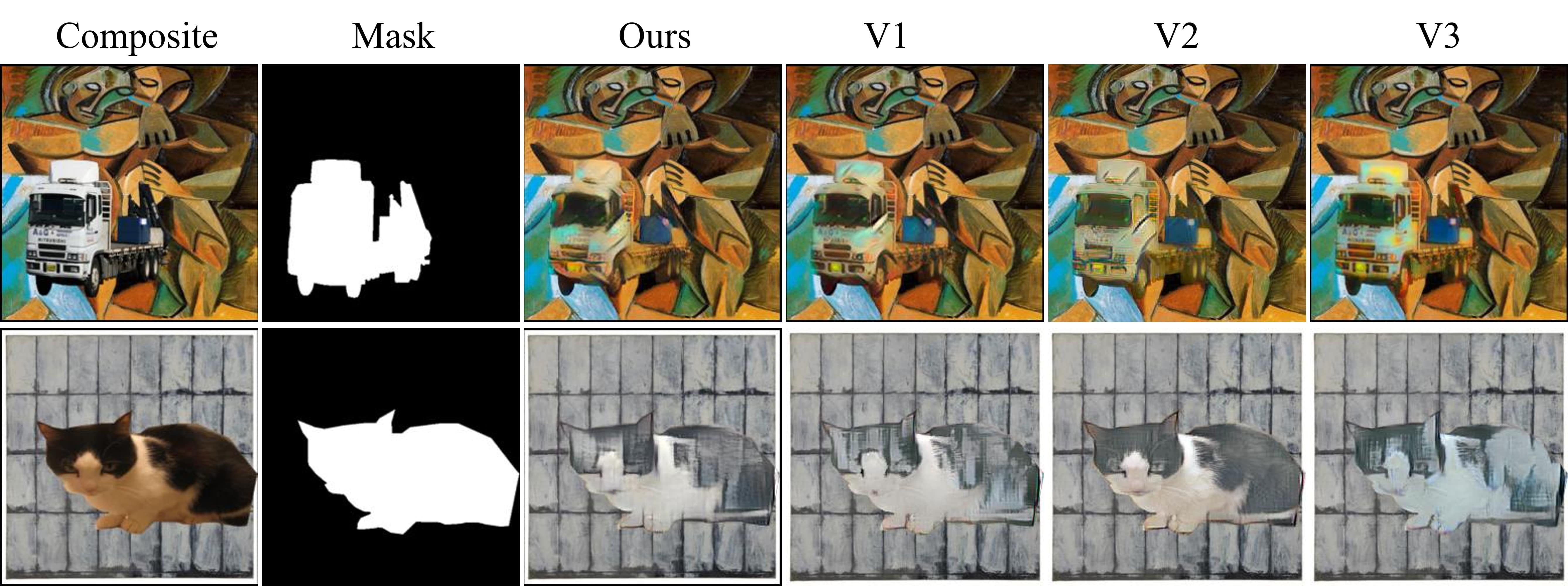}
\caption{From left to right, we show the composite image, the composite mask, the harmonized results of our method and three ablated versions.}
\label{fig:ablation_studies}
\end{figure}

\subsection{Ablation Studies}
Firstly, we try using the full style loss (\emph{i.e.}, $\mathcal{L}^{sty}_4$ in Eqn. \ref{eqn:style_loss}) for all four stages and the harmonized results from four stages are shown in the top row in Figure \ref{fig:full_style_loss}. We also show the harmonized results of our default method for comparison in the bottom row. 
In the top row, even using the full style loss, the outputs ($\tilde{\bm{I}}_1^o$, $\tilde{\bm{I}}_2^o$) from early stages can still only be harmonized up to the low style level. The content structures of outputs ($\tilde{\bm{I}}_3^o$) may even be destroyed when using low-level features for high-level harmonization. 
The results demonstrate that low-level harmonization is simpler than high-level harmonization and high-level features are mandatory for high-level harmonization. 

Secondly, we remove the losses for the first three stages and only use the loss for the final stage. As there is no supervision for the first three stages, we adopt the output from the last stage, which is referred to as $V_1$. 
We also try some variants of our network structure. We replace the upsample decoder with bilinear upsampling, which is referred to as $V_2$. We also reduce the number of fusion  blocks from two to one in each stage, which is referred to as $V_3$. For $V_2$ and $V_3$, we adopt the output from the predicted exit stage. The visualization results of $V_1$, $V_2$, and $V_3$ are shown in Figure~\ref{fig:ablation_studies}.
The harmonized results of $V_1$ become worse than our full method, because the losses in the first three stages can provide useful intermediate supervision and help learn more informative intermediate features. 
The harmonized results of $V_2$ and $V_3$ are also worse than our method, because the simplified network structure reduces the model capacity and degrades the harmonization ability.
The above results verify the necessity of each component in our network.

\section{Conclusion}

In this work, we have developed a progressive harmonization network which can harmonize the composite foreground from low-level styles to high-level styles progressively. We have also designed early-exit strategy, which enables the network to exit at the proper stage. Extensive experiments on the benchmark dataset have demonstrated the superiority of our progressive harmonization network. 

\section*{Acknowledgments}
The work was supported by the National Natural Science Foundation of China (Grant No. 62076162), the Shanghai Municipal Science and Technology Major/Key Project, China (Grant No. 2021SHZDZX0102, Grant No. 20511100300).

\bibliography{main.bbl}

\end{document}


\maketitle

In this document, we provide additional materials to support our main paper. 
In Section~\ref{sec:exit_stage_annotation}, we provide the details and examples for annotated exit stages in the training set. In Section~\ref{sec:visualization}, we provide more visual comparison with baseline methods. In Section~\ref{sec:failure_case}, we discuss the failure cases of our method.

\section{Exit Stage Annotation} \label{sec:exit_stage_annotation}

As introduced in the main paper (Section 3.2 and 4.2 in the main paper),  we randomly select $5000$ composite images in the training set and get their harmonized results from four stages. Then, we ask human annotators to manually select the exit stage for each composite image, in which the exit stage means the earliest stage with satisfactory harmonized output. 
In detail, if the result of one stage is significantly better than the others, they choose this stage as the exit stage. Otherwise, they choose the earliest one from multiple comparable stages. 
Considering the subjectiveness of this annotation task, we ask $50$ human annotators to annotate each composite image and decide the ground-truth exit stage according to major voting, that is, we choose the stage selected by most annotators as the exit stage. The percentages that four stages are annotated as the exit stage are $2.1\%$, $22.3\%$, $47.4\%$, and $28.2\%$ respectively. Overall, the outputs from late stages (\emph{e.g.}, 3, 4) are better than those from early stages (\emph{e.g.}, 1, 2), because the outputs from late stages are usually more sufficiently harmonized in terms of high-level styles. 

\begin{figure}[t]
\centering
\includegraphics[width=0.99\linewidth]{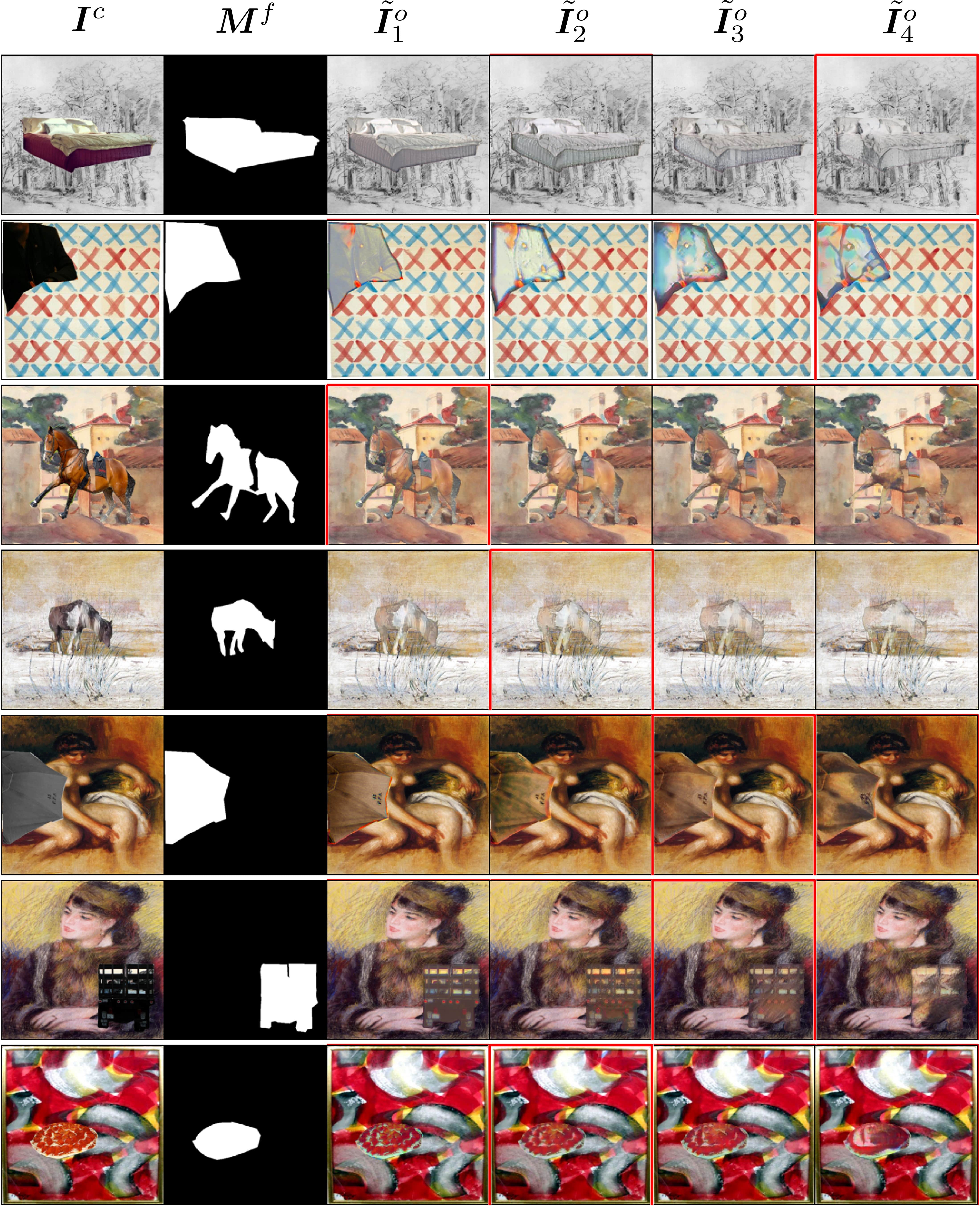}
\caption{From left to right, we show the composite image, composite mask, the harmonized results from four stages in the training set. The annotated exit stages are marked with red bounding boxes.}
\label{fig:exit_stages_train}
\end{figure}

In Figure \ref{fig:exit_stages_train}, we show several examples of composite images and their harmonized results from four stages. We mark the annotated exit stages in red bounding boxes. 
As shown in Figure~\ref{fig:exit_stages_train}, in row 1-2, the results from the first three stages are not sufficiently harmonized, so the last stage is annotated as the exit stage. In row 3-4,  the results from the first two stages are already satisfactory, so the first stage or the second stage are annotated as the exit stages. In row 5-7, the results from the last stage have undesired artifacts (row 5) or distorted content structure (row 6-7), so the stages before the artifacts or distortion appear are annotated as the exit stages. Based on the above observations, our designed early-exit strategy is reasonable because different composite images should exit the network from different stages. Our annotated exit stages on the training set could provide effective supervision for learning to decide the exit stage automatically. 

\begin{figure}[t]
\centering
\includegraphics[width=0.85\linewidth]{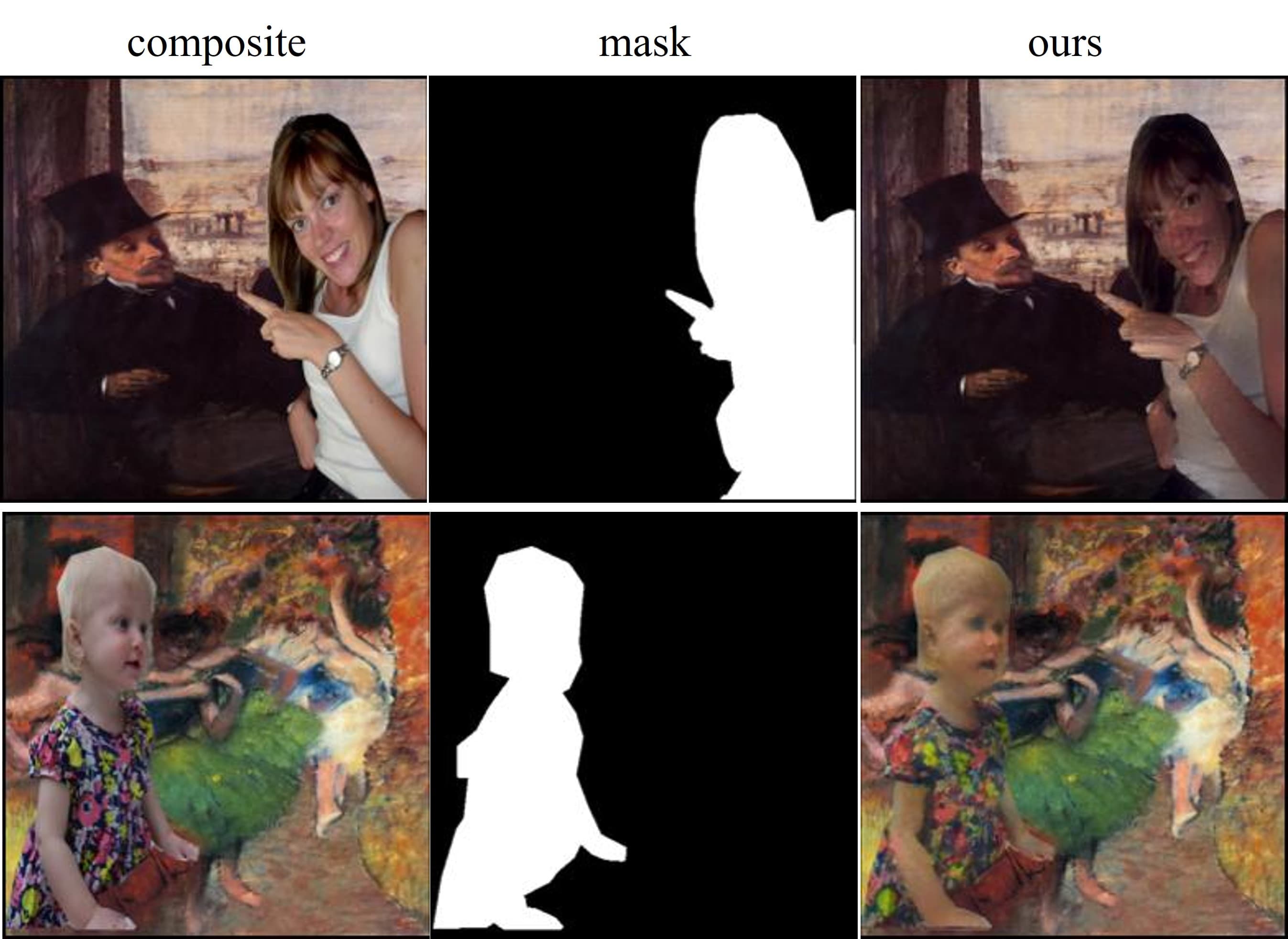}
\caption{Two example failure cases of our ProPIH.}
\label{fig:failure_cases}
\end{figure}

\begin{figure*}[t]
\centering
\includegraphics[width=0.99\linewidth]{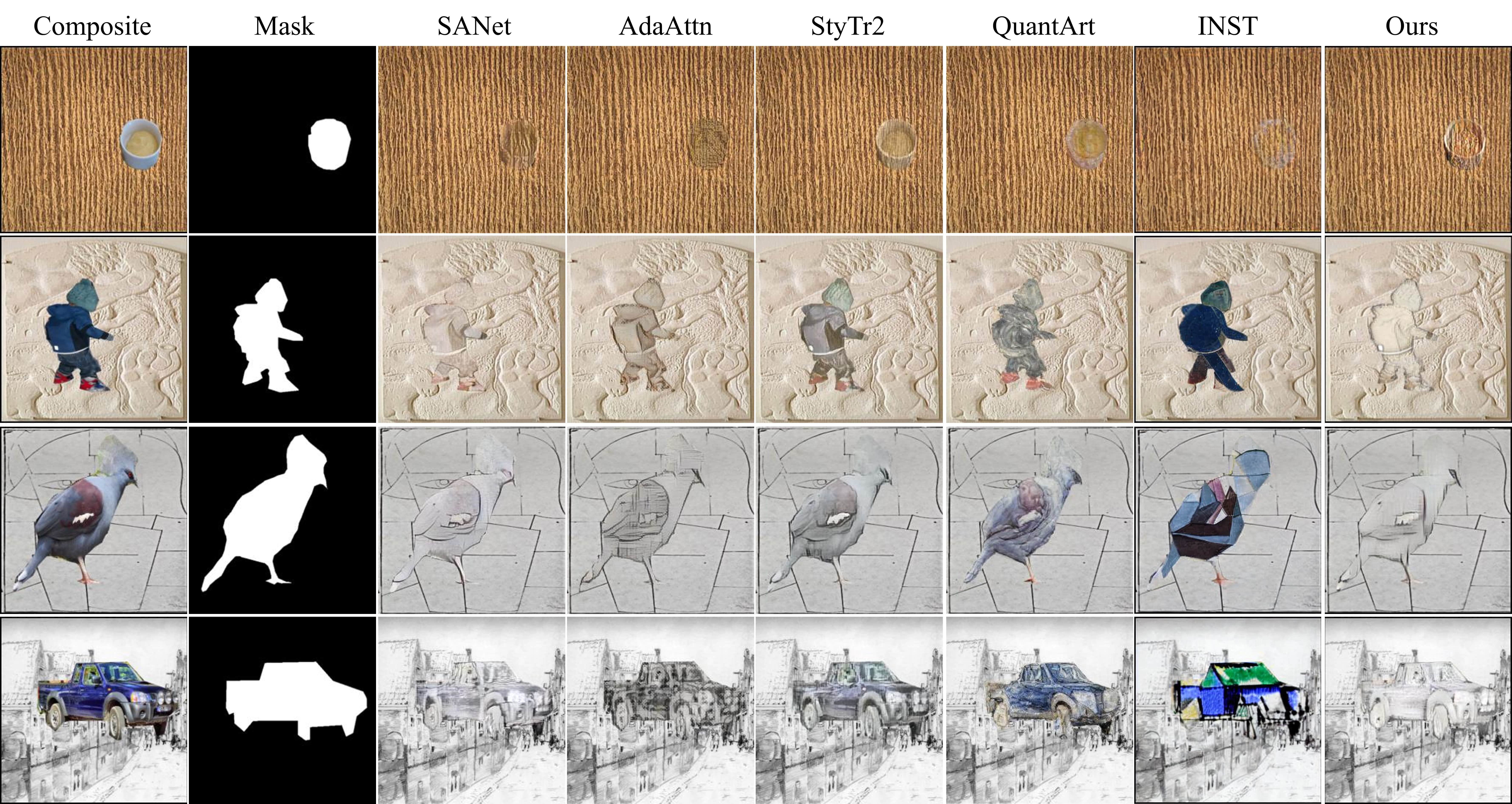}
\caption{From left to right, we show the composite image, composite foreground mask, the harmonized results of  SANet~\cite{park2019arbitrary}, AdaAttN~\cite{liu2021adaattn}, StyTr2~\cite{deng2022stytr2}, QuantArt~\cite{quantart}, INST~\cite{inst}, and our ProPIH.}
\label{fig:style_transfer_baseline_supp}
\end{figure*}

\begin{figure*}[t]
\centering
\includegraphics[width=0.99\linewidth]{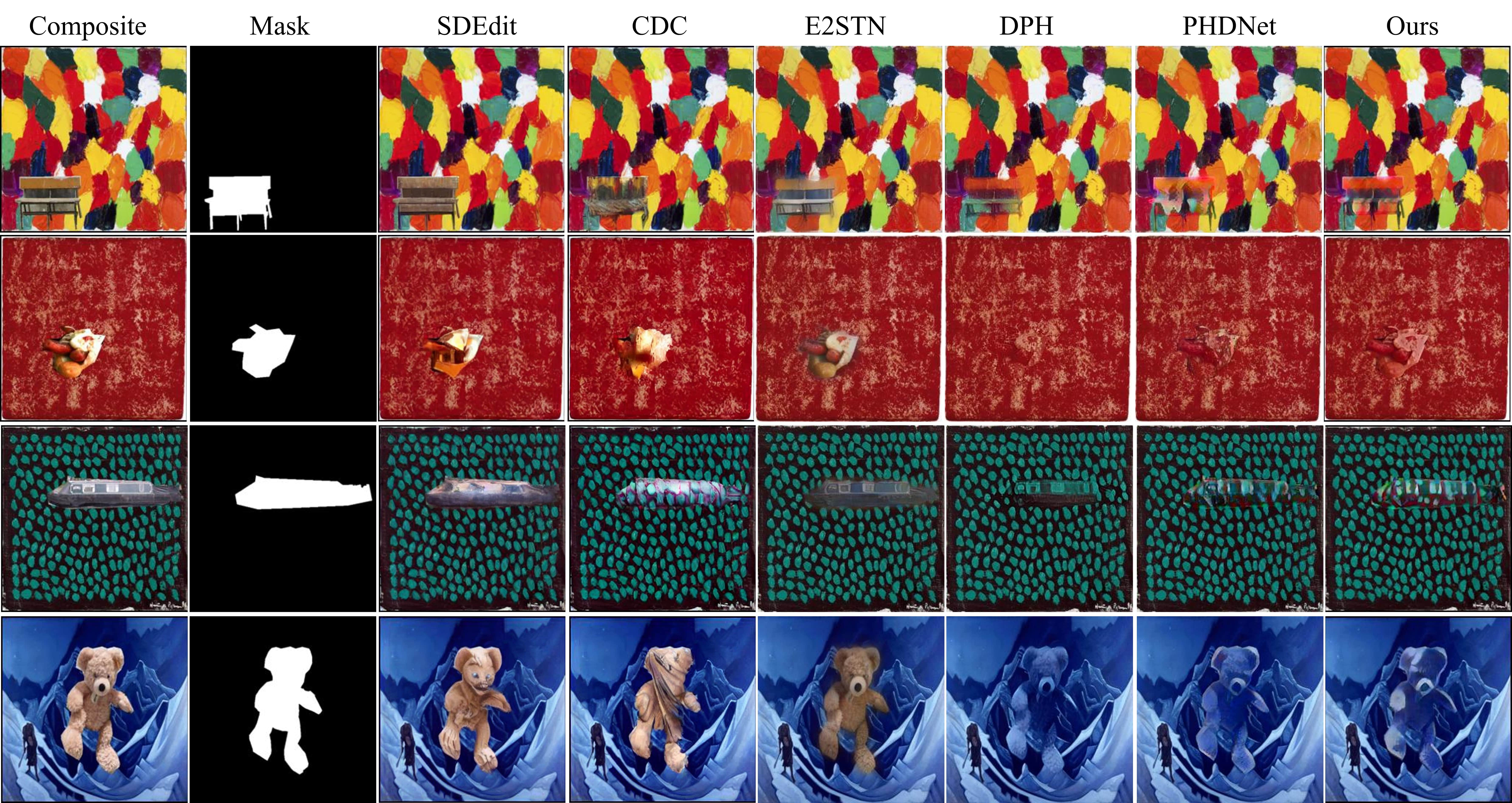}
\caption{From left to right, we show the composite image, composite foreground mask, the harmonized results of SDEdit~\cite{sdedit}, CDC~\cite{cdc}, E2STN~\cite{peng2019element}, DPH~\cite{luan2018deep}, PHDNet~\cite{cao2022painterly}, and our ProPIH.}
\label{fig:painterly_harmonization_baseline_supp}
\end{figure*}

\section{More Visualization Results} \label{sec:visualization}

In Section 4.3 in the main paper, we compare with two groups of baselines. The first group is artistic style transfer baselines: SANet~\cite{park2019arbitrary}, AdaAttN~\cite{liu2021adaattn},  StyTr2~\cite{deng2022stytr2}, QuantArt~\cite{quantart}, INST~\cite{inst}. 
The second group is painterly image harmonization baselines: SDEdit~\cite{sdedit}, CDC~\cite{cdc}, DPH~\cite{luan2018deep}, E2STN~\cite{peng2019element}, PHDNet~\cite{cao2022painterly}. In this section, we provide more comparison results with these two groups of baselines. 

We show the comparison with the first group of baselines in Figure \ref{fig:style_transfer_baseline_supp}. The baselines tend to insufficiently or incorrectly stylize the foreground objects, so that the foreground colors/textures are apparently inconsistent with the background colors/textures (\emph{e.g.}, the boy in row 2, the car in row 4). In contrast, our stylized foregrounds are more compatible with the backgrounds and naturally embedded into the backgrounds.

We show the comparison with the second group of baselines in Figure \ref{fig:painterly_harmonization_baseline_supp}. It can be seen that the baseline methods are struggling to balance stylization and content preservation, so that noticeable artifacts and blurred content are visible in the composite foregrounds. In comparison, our method is able to achieve a good balance between stylization and content preservation, producing visually pleasing results with well-preserved content structures.

\section{Failure Cases} \label{sec:failure_case}

Although our ProPIH can usually achieve satisfactory results, there also exist some failure cases. For example, as shown in Figure~\ref{fig:failure_cases}, when the background is oil painting and the foreground is human portrait, the harmonized human face still looks photorealistic and does not match the background style. One possible reason is that we are very sensitive to the subtle details in human faces, but the network has difficulty in transferring the background style to the subtle facial features.

\section*{Acknowledgments}
The work was supported by the National Natural Science Foundation of China (Grant No. 62076162), the Shanghai Municipal Science and Technology Major/Key Project, China (Grant No. 2021SHZDZX0102, Grant No. 20511100300).

\bibliography{supp.bbl}